%% file: main.tex
\documentclass{article}
\usepackage{booktabs} 
\usepackage{authblk}
\usepackage{multirow}
\usepackage{graphicx}
\usepackage{tabularx}
\graphicspath{ {figures/} }
\newcommand{\eat}[1]{}
\usepackage{listings}
\usepackage{xcolor}
\usepackage{paralist}
\usepackage{makecell}
\usepackage{subfig}
\usepackage{multicol}
\usepackage{float}
\usepackage{url}
\usepackage[all=normal,leading,paragraphs,margins]{savetrees}

\newcommand{\sbjcomment}[1]{\textcolor{blue}{(sbj: #1)}}

\title{DataVizard: Recommending Visual Presentations for
  Structured Data}

\author{Rema Ananthanarayanan}
\author{Pranay K Lohia}
\author{Srikanta Bedathur}
\affil{IBM Research, India}
\begin{document}
\maketitle
\begin{abstract}
Selecting the appropriate visual presentation of the data such that it
preserves the semantics of the underlying data and at the same time
provides an intuitive summary of the data is an important, often the
final step of data analytics. Unfortunately, this is also a step
involving significant human effort starting from selection of groups
of columns in the structured results from analytics stages, to the
selection of right visualization by experimenting with various
alternatives. In this paper, we describe our \emph{DataVizard} system
aimed at reducing this overhead by automatically recommending the most
appropriate visual presentation for the structured
result. Specifically, we consider the following two scenarios: first,
when one needs to visualize the results of a structured query such as
SQL; and the second, when one has acquired a data table with an
associated short description (e.g., tables from the Web). Using a
corpus of real-world database queries (and their results) and a number
of statistical tables crawled from the Web, we show that DataVizard is
capable of recommending visual presentations with high accuracy. 
We also present the results of a user survey that we conducted in order to assess user views of the suitability of the presented charts vis-a-vis the plain text captions of the data.
\end{abstract}

%
%


\maketitle

\input{introduction}

\input{background}
\input{sqlQueries}

\input{noSQL}

\input{architecture}
\input{experiments}

\input{futureWork}

\end{document}

%% file: introduction.tex

\section{Introduction}
\label{introduction}

Visual analytics today is an integral part of the data analytics pipeline and follows as a natural sequel or complement to other modes of data analytics through browsing, querying or search interfaces. Visual presentation of structured data in various formats ranging from simple bar-/line-charts to the use of stacked/grouped bar-charts for multi-dimensional data, facilitate visual analytics of the data by domain experts\footnote{In this paper we use the term chart to refer to the various data representations of structured data such as bar charts, line charts, multi-line charts, stacked bars and pie charts, among others.}. 
In most scenarios, the person who develops these visual presentations, or charts, is skilled in one or more of the domains including data science, statistics, programming and/or user interfaces
.
Given this scenario, the path to data visualization has taken one of the following alternatives:
(i) the visualization is developed as a specific solution for the problem or the domain at hand. 
  This usually pre-supposes domain knowledge and/or extensive knowledge of how the data is to be interpreted as well as how the visualization is going to be used; 
(ii) Standard templates are used, where specific types of data sets result in charts of specific types, and the user then has the option of trying out other chart configurations for the same data. This is the approach in some of the common and popularly used tools like Excel~\cite{msexcel} and Google Sheets~\cite{googlesheets}, where users can highlight the set of rows and columns, view the system-suggested default presentations for the data, and also have the option to view other configurations. Unfortunately, while these systems are quite powerful in terms of repertoire of analysis methods they embed, the support for visualization they extend is quite limited. The suggested visualizations are sensitive to the order of selected columns, and more over, they fail to make use of any metadata -- often available in the form of captions, queries, headers and other forms  -- to appropriately recommend the visualization.

Recently, some attempts have been made to recommend charts for a dataset under consideration by taking natural language hints --- specifically the action clauses in the natural language (such as ``compare'', ``changed over'', etc.)~\cite{articulate}. 
In our paper also we use the natural language semantics to infer the best type of chart. However, in contrast to using the action clauses in the title or query, we focus on automatically generating recommendations by analysing the structured query outputs and Web tables in addition to the types of noun phrases in the title \emph{without} having the user explicitly provide additional natural language hints. 


We have been motivated by the need to be able to recommend the right chart in various scenarios where users query data for insights, either while interacting with a relational database system using  $SQL$ queries or while analyzing the tabular data extracted from the Web or other sources. In these settings, there is often supporting information such as the database query and schema, or the caption of the table, for instance. We started out by building a rule-based visual presentation recommendation system by leveraging the best practices laid out through the  rich work in the area of visual statistics. 
Subsequently, with the insight gained in the process, we were able to define features on the data sets that enabled us to automatically learn the most appropriate charts. Based on this, in DataVizard, which we describe in this paper, we focus on recommending the most relevant chart for the data under consideration. This may then be combined with other visualization generation libraries or tools, to render visual summaries of the data from which the user may get insights for subsequent exploration. Our approach towards achieving this has been two-fold: first, identify the \emph{variables of interest} and the \emph{dependencies} between the variable in a dataset, and, second, recommend the best chart based on the features identified in the first step. 

In the caes of SQL, we consider the variables involved in the query as well as their inter-relationships using information such as the result set and the schema information. In the case of non-SQL data --such as structured statistical tables available on the Web-- we used the caption as the input text describing the data in the same manner as a query. We extract the variables and their dependencies from the caption using standard NLP techniques and combine it with the data in the table to recommend the appropriate chart. 


Overall, the primary contributions of this paper are as follows:
\begin{enumerate}
\item 
We present DataVizard, a system to automatically recommend best visualisations for a structured dataset by taking into consideration the variables of interest and their interdependencies in both SQL as well as non-SQL settings.
\item 
We present the results of our experiments with a couple of SQL workloads over large relational datasets as well as more than 550 non-SQL structured statistical tables crawled from the Web.
\item 
We present the results of a user-survey that we conducted to understand the expectations on the appropriateness of the chart for different queries.
\end{enumerate}

The rest of the paper is structured as follows. In section \ref{background} we discuss related work. 
In sections \ref{SQLsets} and \ref{nonSQL} we describe our technique for finding the variables of interest and recommending the best chart, for SQL data and non-SQL tabular or structured data respectively.
In section \ref{architecture} we describe the overview of our system, and present the system architecture we used for building the recommender. 
In section \ref{experiments}, we discuss our experiments and results obtained for our chart recommendation system and also discuss the results of a user survey on chart preferences. We  conclude in section \ref{futurework} with the future directions and open lines of work.


%% file: background.tex
\section{Background and related work}
\label{background}
Many different streams of work have been actively pursued in the area of data visualization, in different contexts and with different requirements. In this section we discuss some of the representative works.

\begin{figure}[tbh]
\centering
\subfloat[Initial Order]{
  \includegraphics[width=0.7\columnwidth]{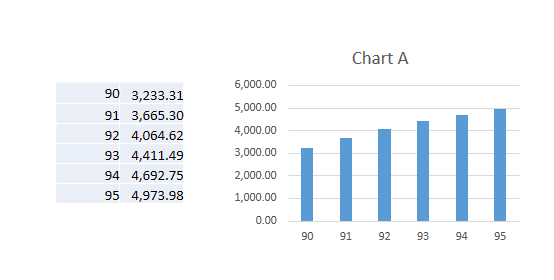}
}

\subfloat[Flipped Columns]{
  \includegraphics[width=0.7\columnwidth]{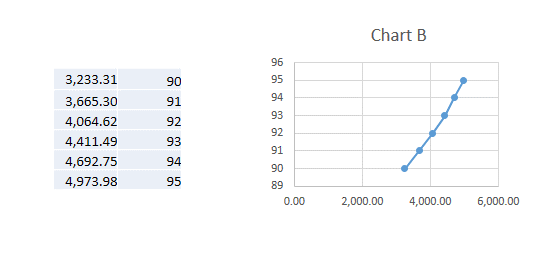}
}
\caption{Issues in Visualization Recommendations for a Portion of the Table \textsf{Public Debt of the United States from 1990 to 2016 (in billions USD)} from MS-Excel}
\label{fig:excel}
\end{figure}

Visual presentation of quantitative data has been extensively studied in the past couple of decades~\cite{cleveland},\cite{tufte}.
Best practices exist that recommend the most appropriate chart for a given data set and a given analytical task, a concise summary of which may be seen in for instance~\cite{abela_fig}. In the business domain, spreadsheets have evolved from just a static representation of data as rows and columns, to highly interactive forms of content. For instance, both MS Excel \cite{msexcel} and Google spreadsheets \cite{googlesheets} allow the user to select a subset of cells, and view the data in these cells as different charts. In simpler cases, the applications recommend relevant charts, based on rules similar to those described in~\cite{abela_fig}, while in others, the user can specify how the data is to be interpreted, and the application generates the appropriate charts on the fly. As we point out in the introduction, these tools, though powerful are limited in their support for recommending visualization. A simple reorganization of the table shown in Figure~\ref{fig:excel} such as swapping the order of columns can result the recommendation switch to a very different visualization (e.g., from bar-chart to line-chart) although it is clear from the context that the \emph{independent variable} that should be in the x-axis is now in the second column.

Tableau \cite{tableau} supports data visualization in a more comprehensive manner, where the user can connect to different data sources like JSON, spreadsheets and text files among others and create a workbook. The user can then select the subset of data to be analyzed. Tableau suggests multiple options for interpreting the data. Further, the user may specify the variables for the axis and the system recommends charts appropriately.  On hovering over different chart options, the system also recommends what needs to be modified in the data to be able to see these types of charts. A key aspect of Tableau is VizQL\cite{vizql} a specification language that describes the structure of a view and the queries used to populate that structure. ShowMe \cite{showme} builds over \cite{vizql} to automatically present data as small sets of multiple views, focusing on the user experience.
VizRec \cite{vizrec} describes the authors`  vision of what visualization recommender systems should have, for identifying and interactively recommending visualizations for a task. 

Another active area of work is in the development of tools like d3\cite{d3} which is a library for creating dynamic interactive charts based on the chart type selected by the user. 

We have seen only a few earlier works that have looked at recommending the best charts for the data in an automated manner. Articulate \cite{articulate} describes a semi-automated model for translating natural language queries into meaningful visualizations. The focus is to provide a streamlined experience to non-expert users. Different types of words in the query express different dependencies and these are then used to derive or suggest a suitable chart. This matches our method closest in terms of the end-goal. However the approach is widely different.In~cite{articulate} it is assumed that users interact with the system through user queries. The queries are parsed to identify features such as comparison\_keywords, relationship\_keywords, composition\_keywords and so on. The keywords for each feature are selected empirically using a bag-of-words model. This differs significantly from our approach where we identify the noun phrases and their types and then try to learn how they are related. Voyager \cite{voyager} supports faceted browsing of recommended charts chosen according to statistical and perceptual measures. It presents a browsing interface that allows interactive browsing of suggested views and aims to balance automation and manual specification. 
In \cite{towardsvizrec} the authors describe a semi-automated visualization recommendation system where information is extracted from the data and the metadata, the latter providing the context. Domain-specific annotations are used on the information, to recommend the visualizations and the system assumes the availability of a knowledgebase of visualizations used in the domain. 

SeeDB\cite{seedb} uses the user query to analyse the space of visualizations and recommends visualizations that appear promising in terms of ``usefulness" or ``interestingness." This is especially useful in high dimensional data, where performance considerations would also be imperative. 

In our work, we specifically focus on automatically recommending the most suitable charts for the given data, based on existing best practices such as those summarised in~\cite{abela_fig}, using both the data and the metadata, which may incluse context information like data captions. No domain knowledge or end-user expertise is assumed and it appears to be one of the first end-to-end fully automated systems for chart recommendations. 
We describe our analysis, observations and experiments in the following sections.


%% file: sqlQueries.tex
\section{Analysis of SQL resultsets}
\label{SQLsets}

The high-level steps in the analysis of both $SQL$ queries and resultsets and non-$SQL$ data comprises $2$ steps a) determining the variables of interest and b) determining their dependency, in terms of the independent variable(s) and dependent variable(s).
In this section we discuss in detail the steps for analysing $SQL$ data and recommending visualizations.

\subsection{Determining the variables of interest }
\label{findingvariablesSQL}
The variables that are related need to be identified from the resultset.
In the most general case, when no information is available except for a list of column names queried, we perform a pairwise identification of columns that may be related. 
Additional cues in the query such as GROUP BY clauses, presence of aggregate functions such as MAX, MIN and others help in identifying the exact relationship between the columns queried. 
Subsequently, where relevant, some pairs that have common variables may be combined into a single visualization.
Further, we categorize each column of data or each variable as either an independent variable, or a dependent variable. Independent variables are those variables that are in many cases categorical in nature. For instance, in a $SQL$ query that selects $employee\_department$ and $average\_income$, $average\_bonus$ for each department, the independent variable is the $employee\_department$, while the dependent variables are the $average\_income$ and the $average\_bonus$. In the most general case, a relationship between $average\_income$ and $average\_bonus$ is not ruled out; however in this specific case, based on the query, we would identify $2$ relationships of interest, viz., $<$$employee\_department$, $average\_income$$>$, and \\$<$$employee\_department$, $average\_bonus$$>$. In the recommendation stage these may be combined in $1$ chart, but at the level of finding variables, these are identified as $2$ relationships. Apart from categorical values, values representing time-periods are in many cases independent variables, for instance, values of some quantities every year or every month.

In the case of $SQL$ queries, a column that is a primary key, or a column that is the predicate of a GROUP BY is an independent variable. 
Some queries may have multiple GROUP BY fields, in which case there are multiple independent variables, with the possibility of creating a grouped chart.
Similarly, a column whose value is aggregated in a query, is generally a dependent variable. By aggregation, here we mean $SQL$ aggregation functions such as $MAX$, $MIN$, $AVG$ and similar.
In the very general case, when no primary key information is available, or when the query has no GROUP BY clause, we will use the methods described in section \ref{nonSQL} to understand the relationship between the terms in the query.
In the case of $SQL$, the number of variables selected is read off the query result. Additional information of the data types of each of the columns can be obtained by querying the schema, while the number of GROUP BY clauses can be obtained by using a $SQL$ parser.

Based on these heuristics, at the end of this analysis, we can tag each column as an independent or dependent variable. For any columns where this is not possible, we fall back on the more general analysis that we descibe in section~\ref{nonSQL}.
\subsection{Recommending the appropriate chart, for $SQL$ resultsets} 
\label{recommendingSQL}
As we mentioned earlier, many best practices exist that suggest which charts are the most appropriate for each data set in terms of what people would like to do with the data --- compare values, analyze trends, see the distributions or compositions. 

In the case of $SQL$ queries, we can capture this intent by analyzing the kinds of variables selected. Further, the actual relationships between the variables, in terms of how we would like to present the data, is determined by the type of the dependent and independent variables. These $2$ items of information suffice to suggest one or more of the relevant visualizations, as depicted in a summary sheet like~\cite{abela_fig}.  For instance, when the independent variable is of string type, as in ``GROUP BY REGION'' or ``GROUP BY DEPARTMENT'' and the dependent variable is a numerical quantity, then we interpret this as a comparison of a numeric quantity across multiple categories, and hence our system would recommend a bar chart. When there is more than one GROUP BY clause, for instance, GROUP BY DEPARTMENT, REGION then our system would recommend a grouped bar chart. When the GROUP BY is related to a time-period variable, as in GROUP BY YEAR or GROUP BY MONTH, then the typical expectation is the trend of a quantity over time, and hence a line chart is shown.
Similarly, when there are multiple GROUP BY clauses relating to time-periods, the system would recommend a multi-line chart.

\subsection{Feature identification for SQL queries}

Sections \ref{findingvariablesSQL} and \ref{recommendingSQL} describe the heuristics we used to determine the most relevant chart for the result sets from a $SQL$ query. 
Based on our learning from the performance of these heuristics (we describe the performance in detail in Section \ref{experiments} )  we identified various features from the $SQL$ queries and the result sets which we then used to train a chart recommendation system, for the $SQL$ queries.  We designed a total of $21$ features which are listed in Table~\ref{tab:sqlfeatures}.

\begin{table}[tbh]
	\footnotesize
\begin{tabular}{l}
\toprule
\bf Features \\
\midrule
select aggregate(A)\\
select B, group by B and B is not time-related field\\
select B,C group by B,C and B and C are time-related\\
select  B, C group by B,C and neither B or C are time-related\\
select  B,C group by B,C and B is time-related and C is not\\
select  B,C group by B,C and B is not time-related and C is \\
select B where B is time-related\\
select B where B is not time-related\\
select A, A is numeric\\
select B, where B is class label\\
select B, where B is primary key or values are unique\\
Presence of group by\\
Presence of multiple group by\\
result set size is $0$\\
result set size is 0 $<$ x $<$= 2\\
result set size is 2 $<$ x $<$= 8\\
result set size is 8 $<$ x $<$=30\\
result set size is x $>$ 30\\
Only $1$ column is selected\\
Exactly $2$ columns are selected\\
More than $2$ columns are selected\\
\bottomrule
\end{tabular}
\caption{Subset of features used for SQL Queries}
\label{tab:sqlfeatures}
\end{table}
The features in Table~\ref{tab:sqlfeatures} are defined in terms of the patterns and presence of various clauses in a $SQL$ query.
A,B,C,~\ldots represent column names. The term \emph{`select'} indicates the query has the SELECT keyword,  The feature \emph{`presence of group by'} implies the query has a GROUP BY clause, while \emph{`presence of multiple group by'} implies a GROUP BY on more than one column. Further, the size of the result set also has a bearing on the chart recommended and this is reflected in the features based on the size of the result set.
Further, while the features are defined in terms of $2$ or at most $3$ columns selected, in the more general case, we can take a pairwise grouping of columns and find the relationship, or when there are many columns in GROUP BY, we can analyse these multiple columns together. We describe some results on the TPC-H queries, later.
The performance of the recommender, based on the training with these features, is described in section \ref{experiments}.
In the next section we describe in detail the analysis of data to recommend charts in the non-SQL setting.

%% file: noSQL.tex
\section{Analysis of non-SQL tabular data}
\label{nonSQL}
Tabular data extracted from pdf files and other sources comprise a set of rows and columns, along with a caption for the table, and optionally, captions for the individual rows and columns. 
We refer to this as the non-SQL setting, and describe techniques to recommend the appropriate chart, by analysing the captions and where needed, the content of the tables. 
We use the caption of each table as the string containing the dependent and independent variables.
Table \ref{tab:titlesExample} gives examples of some captions.
The captions here are in lieu of the SQL queries for SQL data.
The noun phrases in the caption represent the association that is likely desired to be visually depicted, and hence the types of these are used to determine the chart.
We use a natural language parser to extract the common nouns and construct noun phrases. 
We discuss in detail how we parse and analyse these captions to understand the best visual representation for the query and data set.

In terms of the noun phrases in the caption, there are $3$ different possibilities.
\begin{itemize}
\item The number of noun phrases matches the number of variables in the data. This is the most straightforward case, and the noun phrases are the variables we need.
\item The number of noun phrases in the title is less than the number of variables in the data. In such cases one or more variables are implicitly expressed, and need to be deduced; generally it is the one which represents a numeric quantity in the data.
\item The number of noun phrases is more than the number of variables. In this case, we match the noun phrases with columns in the actual data set, based on the column name and data type,  and eliminate the rest of the noun phrases.
\end{itemize}
\begin{table*}
	\footnotesize
	\begin{tabular}{|l|l|}
		\hline
	    Caption Index & Caption\\
		\hline
	    Caption $1$ & Australia: Leading export partners in 2015\\
	    Caption $2$ & Number of employees of Essilor worldwise in 2015, by region\\
	    Caption $3$ & Number of employees of Essilor worldwise from 2008 to 2015\\
	    Caption $4$ & Mobile operating systems market share worldwide from January 2014 to December 2016\\
		Caption $5$ & Box office revenue of the highest grossing movies in North America in 2016 (in million US dollars)\\
	    Caption $6$ & Market share of record labels in Sweden from Dec 26, 2016 to Jan 1, 2017, by single and album charts\\ 
		\hline
    \end{tabular}	    
	\caption{Sample captions of tabular data}
	\label{tab:titlesExample}
\end{table*}
The noun phrases are categorized as indicating object categories, indicating quantitative variables, or indicating time periods.
Further, similar to the GROUP BY YEAR and GROUP BY MONTH clause in SQL, we have phrases like ``from 2005 to 2015" or ``from Jan 2010 to Dec 2012" that form part of the caption, that indicate that there is a time period involved.
If this is validated in the actual data, then the time span comprises one of the independent variables. (In some cases though the title has the time span phrase, the actual data is aggregated for the entitre time period; hence we mention the need to validate with the data.) Similarly, variables that fit in the object-category type also comprise independent variables. Variables that indicate quantity are dependent variables. For instance, in phrases like {\em ``Market share of browsers"}, ``{\em market share}" is the dependent variable while ``{\em browser}" is the independent variable. Hypernyms \footnote{Hypernyms are words that denote type-of relationship. For example {\em parrot} is a type of {\em bird} and hence {\em bird} is hypernym for parrot.} are typically good candidates for independent variables but we will defer its exploration for effective chart recommendation to future work.
\subsection{Determining the variables of interest}
In the case of non-SQL tabular data, the captions of tables or data sets take the place of natural language equivalents of queries for our purposes. 
The parsing and extraction of the various variables is as under.
\begin{asparadesc}
	\item [Identifying noun phrases in captions] We used the Stanford Natural Language Parser~\cite{stanfordnlp} (Stanford NLP) for identifying common nouns, since these are the variables of interest for us in the caption.
	The words tagged `NN' and `NNS' by Stanford NLP were taken as the common nouns of that title.  
	We needed to identify noun phrases also, which, in the context of our work here, we define as the longest sequence of common nouns occurring together.
	This definition varies from the `NP' tag of the Stanford parser, but this is the definition we needed for our current work. 
	The last noun in the noun phrase was treated as the operative noun or the noun that we tried to qualify as quantitative or categorical.
\item [Identifying nouns with quantitative data] We identified common nouns representing quantitative information as follows. 
	A word2vec model with a pre\-trained Google news corpus ($3$ billion running words) word vector model ($3$ million 300-dimension English word vectors) was used. We used the gensim word2vec model \cite{gensim} for our experiment. We defined a set of words which are synonyms of quantifiability - for instance, words such as value, measure, number, numeric, quantity, total, amount and percent. Using this set, we used the word2vec model to find, for each common noun in the title, its similarity score to one of these words in the set. Of all the common nouns in the title, we used the word which has highest score as the word representing a quantity, in that caption. 
\item [Dealing with temporal context terms] In some captions, besides the $2$ or $3$ variables in the data set, the caption had additional information such as point-in-time phrases. For example, in the caption ``{\em US infant mortality by state in 2016}'', the term ``{\em in 2016}'' qualifies the variables but is not needed for identifying the variables. Hence such terms were pruned from each caption before performing the actual variable identification. The extraction of these phrases was done using a bag-of-words to represent such terms, such as months of the year, year patterns and days of the week. 
\item [Identifying caption subject] The subject or topic of each caption was again identified using the Stanford NLP parser as follows. The parse tree for each caption was obtained, with the corresponding parent-child relationship highlighting the dependencies. The tree was traversed in order, to identify the first noun as the subject. In most cases the subject was highly likely to be one of the variables of interest.
\item [Presence of time-span] Each title was classified into $2$ classes, based on whether it contained time-span information or not. For finding time-span information, presence of patterns like ``from $X$ to $Y$'' where $X$ and $Y$ represent point-in-time terms described above, and minor variations of this pattern were used.
If the timespan information was also part of the data, then it was considered an independent variable. As a counter-example, in the caption ``Movies in the US by genre, from 1995 to 2010", the genre is identified as a categorical independent variable, while the time range is identified as another independent time-related variable. However the data consisted of only $2$ columns of information, the movie genre, and the number of movies in each genre; the content did not match any of the time-related information. Hence this variable was eliminated. and the caption was interpreted to represent only $2$ variables. 
\end{asparadesc}

In the case of SQL, the GROUP BY clauses or the primary keys were the independent variables.
In the case of the table captions, most captions have one or more noun phrases that describe the data set presented. 
Further, some noun phrases are preceded by \emph{by} as in ``by genre" or ``by region".
These patterns, and others such as ``by year", ``by region", ``by brand" and so on, indicate that the data is to be grouped along these categories, similar to the GROUP BY column in $SQL$.
We describe how these heuristics are applied in the examples in~\ref{tab:nonsqlqueries}.
\subsection{Feature identification from caption strings}
In the above paragraphs we have described how we identify the noun phrases of interest and the independent and dependent variables. We now describe how we have used this learning to describe features on the caption strings, also extending from our learning on feature description for $SQL$ queries. These features are then used to train a recommender to recommend the right chart, in the case of tabular data with captions. 

Table \ref{tab:featurestable2} lists the features we have used in the case of non-SQL data sets. The terms used in this table are as follows. $Q$ indicates a quantitative noun, while $T$ indicates a time span expression. $OC$ indicates an object category. ``byT", ``byOC" and ``byQ" indicate the presence of by\- clauses, followed by a time-related variable, an object category type variable and a quantitative variable respectively. For instance, a table caption that states, \emph{``Market share of browsers"} would be treated, after parsing, as ``Main phrase is $Q$ and secondary phrase is $OC$", while a caption that states \emph{``Unemployment rate in Florida from 1992 to 2015"} would be treated as ``Main phrase is $Q$ and secondary phrase is $T$".

\begin{table}
\footnotesize
\begin{tabular}{l}
\toprule
\bf Features \\ 
\midrule
Main phrase is Q, secondary phrase is T \\
Main phrase is Q, secondary phrase is OC \\
Main phrase is T, secondary phrase is Q \\
Main phrase is OC, secondary is Q\\
Main phrase is Q, prepositional phrase is byT \\
Main phrase is Q, prepositional phrase is byOC \\
Main phrase is T, prepositional phrase is byQ \\
Main phrase is OC, preprositional phrase is byQ \\
Main phrase is Q, has timespan phrase \\
Main phrase is Q, has OC1 and OC2\\
Main phrase is OC, has Q1 and Q2\\
Main phrase is Q, has OC and T\\
Main phrase is Q, has T1 and T2\\
Main phrase is OC, has term 'distribution'\\
Only 1 data point \\
Only 2 data points \\
2 $<$ nbr of data points $<$= 8 \\
8 $<$ nbr of data points $<$=30 \\
30 $<$ nbr of data points \\
\bottomrule
\end{tabular}
\caption{ Subset of features used for the captions }
\label{tab:featurestable2}
\end{table}

In the next section we describe our experiments to recommend the most appropriate charts.

%% file: architecture.tex
\section{Architecture of DataVizard}
\label{architecture}

\begin{figure}[tbh]
\includegraphics[width=\columnwidth]{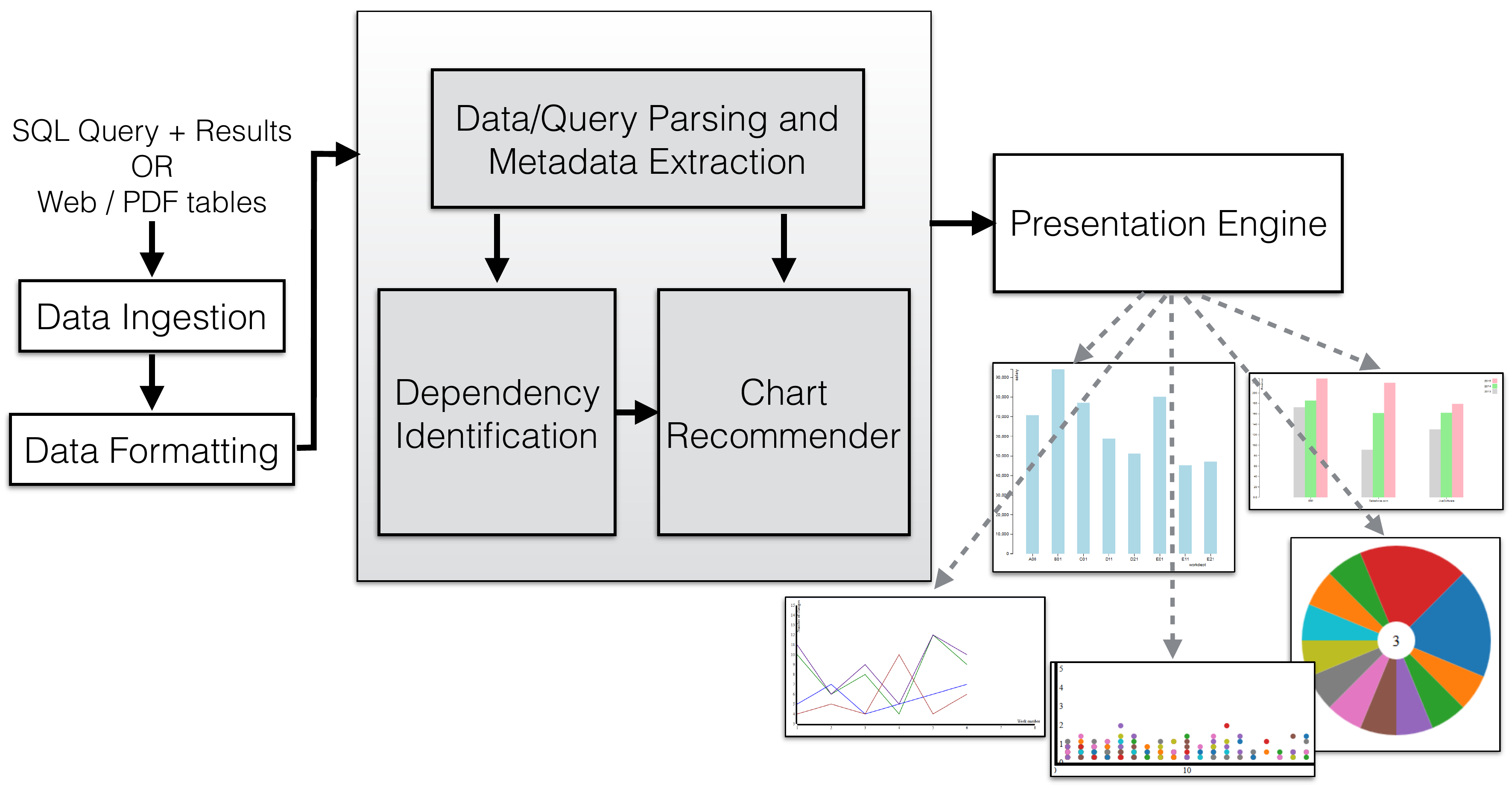}
\caption{Components in the DataVizard System}
\label{arch_fig}
\end{figure}
In this section we describe the overall architecture of DataVizard, our system for recommending and presenting the most appropriate chart for a data set. The end-goal of our solution is to automatically recommend visualizations for complementing the insights from other modes of data analytics in various different scenarios that may include analysis of $SQL$ result sets, analysis of tabular data in spread sheets, tables of information extracted from the web, or tables extracted from PDF files. In the context of this paper we are not concerned with the framework or libraries that may be used to generate the visualization and limit our scope to recommending the best chart for the data presented. Further, while we have used existing NLP tools in parsing and extraction of variables, we have not  focused on improving the extraction efficiency, but on identification of features that help in recommending the best visualization. With this objective, our architecture is designed with the following components. Figure~\ref{arch_fig} shows the high-level view of our architecture. The following are the main components.

\begin{asparadesc}
\item [Data ingestion and formatting:] The input data could be from various sources and formats. This component reads the input data and formats it in JSON, in a pre-defined format. The subsequent stages use this JSON data. The JSON encapsulates the actual data, the metadata information which includes primary and foreign key relationships in the case of SQL, queries and table captions where available and any other metadata that may be available.
\item [Data parsing and metadata extraction:] In the case of SQL, metadata information such as the number of components selected, the data type of each column selected, the presence and number of GROUP BY columns and the presence of various aggregate functions are extracted. This could be done by using any of the standard SQL parsers that are available. In the case of non-SQL data, table captions have information such as noun phrases, object phrases, presence of time span periods and other cues. These details are extracted, which are then analyzed in the subsequent stages.
\item [Dependency identification:] We need to identify the independent and the dependent variables in the relationship represented in the dataset or captured in the query or data caption, in order to recommend the best-fit chart. A typical example of variable dependency is, in a query to display the annual revenue of a company, the independent variable is the year, and the dependent variable is the revenue. SQL data and non-SQL data have distinct cues that help in identifying the independent and dependent variables, as we have described in sections~\ref{SQLsets} and \ref{nonSQL}. 
\item [Recommender:] This component analyses the properties of the independent and dependent variables identified in the earlier step, and factors in the information on the data sizes to map it to relationships defined in existing best practices to recommend the appropriate charts. 
\end{asparadesc}

As we mentioned in section \ref{background}, prior work has attempted to capture the type of relationship between data items by focusing on the non-noun clauses in the query to determine what association is being shown. In our work, we have attempted to determine the association in terms of the variables being depicted. For example, \emph{comparison} of values \emph{over time} and \emph{across categories} are treated as two different associations despite having the same action clause in the query. Further, existing solutions are not able to combine multiple relationships into one chart by doing a logical grouping, unless explicitly selected by the user.  Our method allows us to combine multiple variables, where related, in one visualization. We discuss our results on the heuristics-based approach and the features-trained approach in Section~\ref{experiments}, with examples.

%% file: experiments.tex
\section{Experimental Results}
\label{experiments}

In this section we discuss the results obtained with our recommender system. We also performed a user survey of charts to understand how the implicit assumptions with which users map captions to visualizations matches or compares with the explicit mappings using our methods of identifying the noun phrases, and assigning a type for each. We also discuss the results of our survey.

The initial recommendation engine was built as a rule-based recommendation engine. We focused our attention on the set of charts that are often preferred by business analysts, listed below.
\begin{multicols}{2}
\begin{itemize}
\item Line
\item Horizontal bar
\item Vertical bar 
\item Pie
\item Multi-line
\item Grouped bar
\item Stacked bar
\item Scatter
\item Table
\end{itemize}
\end{multicols}

The initial set of charts for a small set of $SQL$ data sets and the non-$SQL$ data sets were determined based on a set of rules relating the independent variable and the dependent variable. As expected, the accuracy was very high. Also this helped to engineer features for automatically learning the most appropriate chart.
\subsection{Results over SQL Queries and Tables}
We tested our initial heuristics-based recommender on the queries in the TPC-H benchmark data set\cite{tpch} and subsequently extended this to learn features for the $SQL$ queries. From each $SQL$ query, we used the subset of components that we needed to recommend the chart. Some of these components were from the query string itself, and some from the metadata. 
In Figure~\ref{fig:tpchqueries} we list $3$ queries from the TPC-H set and we then describe in detail how we perform the actual analysis. The reason we used the TPC-H query set was, this is a standard set which also contains the patterns that captured the features we have used in our system. 
\begin{figure*}[tbh]
  \begin{minipage}[t]{0.3\textwidth}
\textbf{Query 1:}
\begin{lstlisting}[
           language=SQL,
           showspaces=false,
           basicstyle=\footnotesize\ttfamily,
           frame=l,
           xleftmargin=0pt,
%           numbers=left,
           breaklines=true, numberstyle=\tiny,
           commentstyle=\color{gray} ]
select l_returnflag, l_linestatus, sum(l_quantity) as sum_qty,
sum(l_extendedprice) as sum_base_price, sum(l_extendedprice * (1 -
l_discount)) as sum_disc_price, sum(l_extendedprice * (1 - l_discount)
* (1 + l_tax)) as sum_charge, avg(l_quantity) as avg_qty,
avg(l_extendedprice) as avg_price, avg(l_discount) as avg_disc,
count(*) as count_order from lineitem where l_shipdate <= `1998-12-01`
group by l_returnflag, l_linestatus order by l_returnflag, l_linestatusem
         \end{lstlisting}
       \end{minipage}%
\quad %
\begin{minipage}[t]{0.3\textwidth}
\textbf{Query 4:}
\begin{lstlisting}[
        language=SQL,
        showspaces=false,
        basicstyle=\footnotesize\ttfamily,
           frame=l,
           xleftmargin=0pt,
%           numbers=left,
        breaklines=true, numberstyle=\tiny, commentstyle=\color{gray}
        ]
select o_orderpriority, count(*) as order_count from orders
where o_orderdate $>$= `1993-07-01` and o_orderdate <
DATE(`1993-07-01`) + 3 months and exists ( select * from
lineitem where l_orderkey = o_orderkey and l_commitdate <
l_receiptdate ) group by o_orderpriority order by
o_orderpriority
      \end{lstlisting}
    \end{minipage} %
\quad %
\begin{minipage}[t]{0.3\textwidth}
\textbf{Query 22:} \begin{lstlisting}[
           language=SQL,
           showspaces=false,
           basicstyle=\footnotesize\ttfamily,
           frame=l,
           xleftmargin=0pt,
%           numbers=left,
           breaklines=true,
           numberstyle=\tiny,
           commentstyle=\color{gray}
        ] 
select cntrycode, count($*$) as numcust, sum(c_acctbal) as totacctbal from 
( select substr(c\_phone,1,2) as cntrycode, c_acctbal from customer 
  where substr(c\_phone,1,2) in (`13`, `31`, `23`, `29`, `30`, `18`, `17`) and c_acctbal > 
( select avg(c_acctbal) from customer 
where c_acctbal > 0.00 and substr(c_phone,1,2) in (`13`, `31`, `23`, `29`, `30`, `18`, `17`) ) and not exists ( select * from orders where o_custkey = c_custkey ) ) as custsale 
group by cntrycode order by cntrycode
\end{lstlisting}
\end{minipage}
    \caption{Example Queries from TPC-H Benchmark}
    \label{fig:tpchqueries}
  \end{figure*}

Table \ref{tab:sqlqueries} lists the variables identified by our heuristics and the charts recommended, the query patterns that determine the features and the actual features used for learning, in the subsequent step of automated chart recommendation.
Note that the actual $SQL$ query may be quite long, querying and joining across multiple tables, but the information we need for the chart determination is mainly a subset of the patterns and the result set size.
In Table \ref{tab:sqlqueries} the line labeled ``Pattern" indicates the information used for each SQL query. Through this we identified a total of $21$ features, which were listed in Table~\ref{tab:sqlfeatures}.

\begin{table*}[tbh]
\footnotesize
\begin{tabularx}{\textwidth}{c | c X}
\toprule
\multirow{5}{*}{Query 1} & \textbf{Independent variables:}& \texttt{l\_returnflag, l\_linestatus} -- based on the presence of \texttt{GROUP BY}\\
						&\textbf{Dependent variables:}& 
\texttt{sum\_qty, sum\_base\_price, sum\_disc\_price, sum\_charge, avg\_qty, avg\_price, avg\_disc, count\_order} -- based on the aggregate function use\\
			&\textbf{Recommendation:} & 
8 grouped bar charts, where each bar chart has the x-axis of \texttt{l\_returnflag}, grouped on \texttt{l\_linestatus}\\
&\textbf{Pattern:} & \texttt{Select A, B}; \texttt{AGGREGATE\_FUNCTION(C)}; and \texttt{GROUP BY A,B}\\
&\textbf{Features:} & Select aggregated(A); Select B,C GROUP BY B,C where neither B or C is time-related; 2 $<$ number of rows in result $<$= 8 \\
\midrule
\multirow{5}{*}{Query 4}  & \textbf{Independent variables:} &  \texttt{o\_orderpriority} -- based on the \texttt{GROUP BY}\\
&\textbf{Dependent variables: }& 
           \texttt{order\_count} -- based on the aggregation function \texttt{COUNT}\\
&\textbf{Recommendation: }& A bar chart, with \texttt{o\_orderpriority} on x-axis and \texttt{order\_count }on y-axis.\\
&\textbf{Pattern:}& \texttt{Select A}; \texttt{AGGREGATE\_FUNCTION(B)}; \texttt{GROUP BY A} \\
&\textbf{Features:}& Select aggregated(A); Select B GROUP BY B where B is not time - related; 2 $<$ number of rows in result $<$= 8 \\
\midrule
\multirow{5}{*}{Query 22} & \textbf{Independent variables:}& \texttt{cntrycode} -- based on the \texttt{GROUP BY}\\
&\textbf{Dependent variables:}& \texttt{numcust} -- based on aggregation function \texttt{COUNT} and \texttt{totacctbal} -- based on \texttt{SUM}\\
&\textbf{Recommendation:}& 2 bar charts are recommended, \texttt{cntrycode} vs \texttt{numcust} and \texttt{cntrycode} vs \texttt{tatacctbal}\\
&\textbf{Pattern: }&\texttt{Select A}; \texttt{AGGREGATE\_FUNCTION(B)}; \texttt{AGGREGATE\_FUNCTTION(C)}; \texttt{GROUP BY A} \\
&\textbf{Features:}& 2 sets of features of type: Select aggregated(A); Select B GROUP BY B where B is not time - related; 2 $<$ number of rows in result $<$= 8 \\
\bottomrule
\end{tabularx}
\caption{Features extracted and recommendation}  
\label{tab:sqlqueries}
\end{table*}

\begin{table*}[tbh]
\footnotesize
\begin{tabularx}{\textwidth}{c | c X}
\toprule
\multirow{4}{*}{Caption 1} 
	& \textbf{Noun phrases:} &  \texttt{partners} -- from parser output\\
	& \textbf{Independent variables:} &  \texttt{partners} -- it is the default\\
	&\textbf{Dependent variables: }& \texttt{share} -- it is implicit, and deduced from data\\
	&\textbf{Recommendation: }& A bar chart, with \texttt{partners} on x-axis and \texttt{share }on y-axis.\\
&\textbf{Features:}& Main phrase is OC, prepositional phrase is byQ; 2 $<$ number of rows in result $<$= 8 \\
\midrule
\multirow{4}{*}{Caption 2} 
	& \textbf{Noun phrases:} &  \texttt{employees, region} -- from parser output\\
	& \textbf{Independent variables:} &  \texttt{region} -- based on the term \texttt{by}\\
	&\textbf{Dependent variables: }& \texttt{employees} -- by elimination of other variables\\
	&\textbf{Recommendation: }& A bar chart, with \texttt{region} on x-axis and \texttt{employees (number of) }on y-axis.\\
&\textbf{Features:}& Main phrase is Q, prepositional phrase is byOC; 2 $<$ number of rows in result $<$= 8 \\
\midrule
\multirow{4}{*}{Caption 3} 
	& \textbf{Noun phrases:} &  \texttt{employees} -- from parser output\\
	& \textbf{Independent variables:} &  \texttt{year} -- based on the time span pattern in caption\\
	&\textbf{Dependent variables: }& \texttt{employees} -- by elimination of other variables\\
	&\textbf{Recommendation: }& A line chart, with \texttt{year} on x-axis and \texttt{employees (number of) }on y-axis.\\
&\textbf{Features:}& Main phrase is Q, prepositional phrase is byT; 2 $<$ number of rows in result $<$= 8 \\
\midrule
\multirow{4}{*}{Caption 4} 
	& \textbf{Noun phrases:} &  \texttt{systems, market share} -- from parser output\\
	& \textbf{Independent variables:} &  \texttt{month, systems} -- based on the time span pattern identified, and matched with actual data\\
	&\textbf{Dependent variables: }& \texttt{share} -- quantitative phrase\\
	&\textbf{Recommendation: }& A multi-line chart, with \texttt{month} on x-axis and \texttt{market share }on y-axis, one line graph for each (mobile operating) system.\\
&\textbf{Features:}& Main phrase is Q, prepositional phrase is byT; 8 $<$ number of rows in result $<$= 30 \\
\midrule
\multirow{6}{*}{Caption 5} 
	& \textbf{Noun phrases:} &  \texttt{box office revenue, movies} -- from parser output\\
	& \textbf{Independent variables:} &  \texttt{movies} -- based on elimination of quantitative term\\
	&\textbf{Dependent variables: }& \texttt{(box office) revenue} -- quantitative phrase\\
	&\textbf{Recommendation: }& A horizontal bar chart, with each bar representing a movie and \texttt{revenue }on x-axis\\
&\textbf{Features:}& Main phrase is Q, prepositional phrase is byOC; 8 $<$ number of rows in result $<$= 30 \\
\midrule
\multirow{6}{*}{Caption 6} 
	& \textbf{Noun phrases:} &  \texttt{market share, record labels, album charts } -- from parser output\\
	& \textbf{Independent variables:} &  \texttt{single charts, album charts} -- based on \texttt{by} and mapping with data column size\\
	&\textbf{Dependent variables: }& \texttt{market share} -- quantitative phrase\\
	&\textbf{Recommendation: }& A grouped bar chart, with each group representing a label and one bar each for single chart and album chart \texttt{market share }on x-axis\\
&\textbf{Features:}& Main phrase is Q, prepositional phrase is byOC, byOC; 2 $<$ number of rows in result $<$= 8 \\
\bottomrule
\end{tabularx}
	\caption{Semantic analysis of captions and features}
\label{tab:nonsqlqueries}
\end{table*}

\input{charts}

We then tested our recommender on a dataset of $68$ $SQL$ queries over a financial services knowledge base obtained from the authors of the paper~\cite{athena}. We transformed these queries into the same feature representation over $21$ features obtained from the TPC-H queries earlier. Note that these features are labeled in binary where 1 (0) represents the presence (absence) of a feature. The class labels --i.e., the best recommended chart as well as the second-best recommended chart were manually applied using the rules derived from various best practices models~\cite{abela_fig}.

\begin{table*}[tbh]
\begin{tabular}{l c c c c c c c}
\toprule
\multirow{2}{*}{K-fold}	& \multicolumn{7}{c}{Maximum Features}\\
	\cline{2-8}
 & Auto &21 &16 &14 &10 &8 &4\\
\midrule
10 &  0.9559, 0.9559  & 0.9559, 0.9559 & 0.9411, 0.9558 & 0.9411, 0.9559 & 0.9559, 0.9588 & 0.9559, 0.9588 & 0.9850, 0.9850 \\

5 &  0.9260, 0.8676 & 0.9412, 0.8520 & 0.9411, 0.8676 & 0.9265, 0.8676 & 0.9411, 0.8676 & 0.8382, 0.8676 & 0.8382, 0.8676 \\

3 &  0.9411, 0.8676 & 0.8970, 0.8530 & 0.9411, 0.8676 & 0.8970, 0.8676 & 0.9411, 0.8676 & 0.8970, 0.8676 & 0.8970, 0.8676 \\
\bottomrule                                                                                                                                                            
\end{tabular}
\caption{Accuracy at rank 1 and 2 for the SQL data set}
\label{tab:sqlaccuracy}
\end{table*}

We trained our recommendation model using the random forest classifier~\cite{randomforest2} and performed cross-validation with varying k-fold (10, 5 and 3) as indicated in column $1$ of table~\ref{tab:sqlaccuracy}. The table shows the accuracy for the recommended charts, where the first number in each pair is the accuracy for the first recommended chart and the second number is the accuracy for the second recommended chart. The random forest has one hyper-parameter, i.e., \emph{the maximum number of features} used by the classifier. We ran our experiments for different values of the hyper-parameters including auto, which denotes the square root of the total number of features (i.e., 21),$16$,$14$,$10$,$8$ and $4$.
The very high precision values can be attributed to the fact that most of the queries were homogeneous in nature and were not very complex in terms of the features.

\begin{table*}
\begin{tabular}{lccccccc}
\toprule
\multirow{2}{*}{K-fold}	& \multicolumn{7}{c}{Maximum Features}\\
	\cline{2-8}
& Auto &19 &16 &14 &10 &8 &4\\
\midrule
10 &  0.9726, 0.9781  & 0.9744, 0.9781 & 0.9763, 0.9781 & 0.9744, 0.9781 & 0.9744, 0.9781 & 0.9763, 0.9781 & 0.9708, 0.9744 \\

5 &  0.9763, 0.9763 & 0.9763, 0.9781 & 0.9744, 0.9781 & 0.9763, 0.9781 & 0.9744, 0.9781 & 0.9763, 0.9781 & 0.9744, 0.9763 \\

3 &  0.9670, 0.9690 & 0.9581, 0.9598 & 0.9581, 0.9564 & 0.9563, 0.9564 & 0.9480, 0.9480 & 0.9480, 0.9321 & 0.9321, 0.9321 \\
\bottomrule
                                                                                                                                                            
\end{tabular}
\caption{Accuracy at rank 1 and 2 for the table captions data set}
\label{accuracy_550}
\end{table*}

\subsection{Results on Non-SQL Tables}
In the second part of the experiment we tested the automatic recommendation of charts on textual descriptions of the charts. 
Each chart was described by the caption of the chart. We defined the features for the natural-language strings by extending and extrapolating from the features that we defined for the $SQL$ data sets. 
The data comprised $550$ tables downloaded from the web as csv (Comma-Separated Value) files from the site \cite{statista}. This site serves statistical data on a whole variety of topics, and all the data we have used is non-proprietary and not of restricted availability. The site facilitated the access of all the data from a single point, in a curated and cleansed format. The tables were selected at random, covering a wide range of topics of which table \ref{tab:titlesExample} is a small sample. The captions to the tables were used as surrogates for the natural language query and we analyzed these captions, to obtain the component parts of the query. The steps in the identification of the components were performed as described in section~\ref{architecture}. We identified a set of $19$ features, which were described in Table~\ref{tab:featurestable2}.

We compared the performance of our heuristics-based feature extraction with the hand-labeled set for the 550 captions. We got a recall of 75\% and a precision of 90\%. Our heuristics were able to extract around 75\% of the needed variables.  One reason for the lower recall was that in some cases, the caption was very succinct, and the quantitative variable was implicit. This was especially the case when the tabular information was reporting survey results. For a heading like \emph{``Primary attitude towards shopping on social media sites"} which was essentially presenting a survey result, we need to add additional heuristics to extract  or extrapolate the implicit dependent variable  automatically. The variables extracted were mostly correct. Since our focus was on the identification of features for learning to recommend visualizations, we have not investigated further on improving the extraction accuracy here.  We have subsequently used the hand-labeled set as input for training the recommender.

Each caption was then labelled with the features, where a $1$ indicates the presence of a feature and a $0$ indicates its absence, in the caption. In order to get a reference labelled set, we performed a controlled user study with a small set of users on a subset of the data to generate the reference labels. These were users well-versed in visualization best practices. $3$ captions were identified for each representative pattern from the full set of captions available. The set of possible charts for each of these captions was manually generated and $3$ users were required to label each of the selections as ``Very good", ``Good", ``Maybe ok" and 	``Poor". These were then compiled and the first and second class labels were chosen based on the ranking from the user survey. These recommendations were then extrapolated to all the $550$ charts and the random forest classifier was again run on this data set, similar to the case in $SQL$ data. The random forest classifier was run with varying maximum features and cross-validation with varying k\-fold. The predicted class labels were compared with the original class labels obtained by extrapolating our user survey results and the accuracy figures are reported in the table \ref{accuracy_550}. The entries in this table are interpreted similar to the entries in Table~\ref{tab:sqlaccuracy}.
In Table~\ref{tab:nonsqlqueries} we describe the steps in detail for the sample captions listed in Table~\ref{tab:titlesExample}. Figure~\ref{fig:sampleCharts} shows the charts recommended and generated for the captions in Table~\ref{tab:titlesExample}.
\subsection{Note on the classification method}
Decision trees usually come in mind as a first choice when we think of classification. However, the decision tree is a single tree whereas random forest is an ensemble of many decision trees. Further, our data points are limited to a few hundreds, and random forest will perform better in this context.\cite{randomforest1}.
Random forest are easier to tune as there is only one hyperparameter, the maximum number of features. Generally, the maximum number of features for random forest is chosen as the square root of the number of total features. However, varying it along with the K-fold for cross validation gives a bigger picture of the feature list and classification. Other methods like Gradient Boosted Decision Trees are difficult to tune as they have several hyperparameters and are also prone to overfitting for this size of the data.
Although random forest are also likely to overfit the data, they are more robust because tuning the lone hyperparameter solves the problem. 

\input{survey2}


%% file: charts.tex

\begin{figure*}[ht]%
\centering
	\subfloat[Australia: Leading exposty partners in 2015]{{\includegraphics[width=0.25\columnwidth]{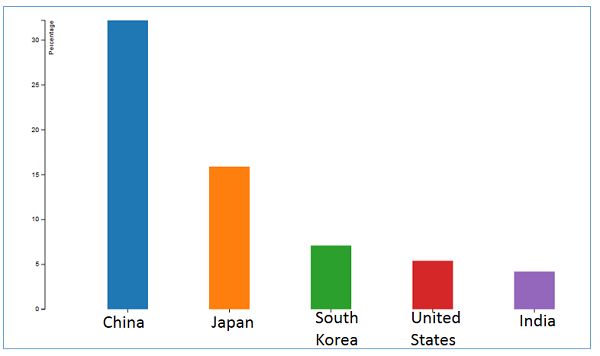}}}%
	\qquad
	\subfloat[Number of employees of Essilor worldwide, in 2015, by region]{{\includegraphics[width=0.25\columnwidth]{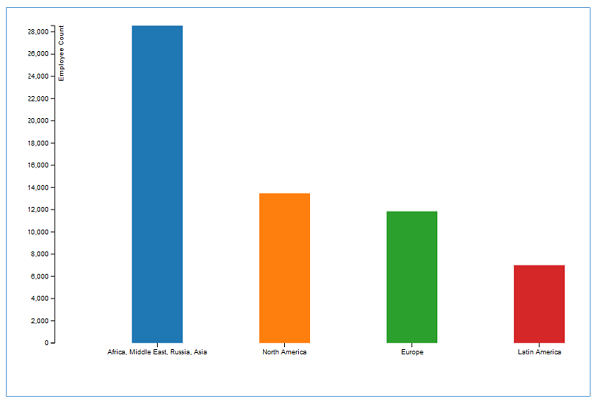}}}%
	\qquad
	\subfloat[Number of employees of Essilor worldwide, from 2008 to 2015]{{\includegraphics[width=0.25\columnwidth]{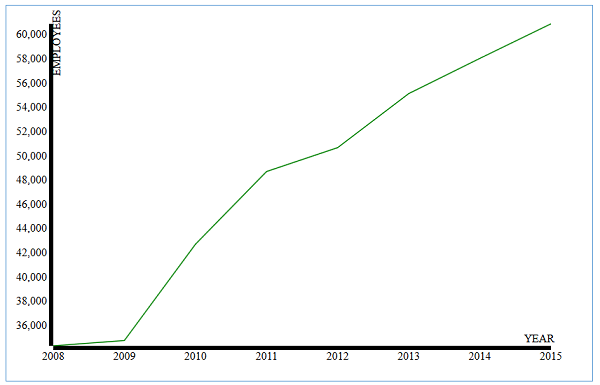}}}%
	\qquad
	\subfloat[Mobile operating systems market share worldwide from Jan 2014 to Dec 2016]{{\includegraphics[width=0.25\columnwidth]{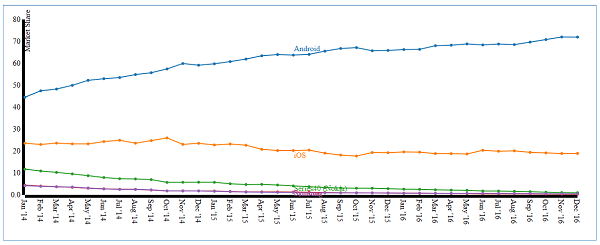}}}%
	\qquad
	\subfloat[Box office revenue of the highest grossing movies in North America in 2016 (in million USD)]{{\includegraphics[width=0.25\columnwidth]{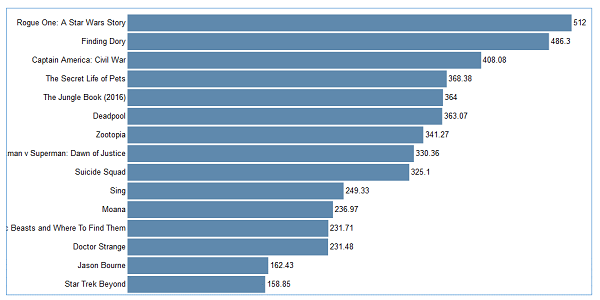}}}%
	\qquad
	\subfloat[Market share of record labels in Sweden from Dec 26, 2016 to Jan 1, 2017, by single and album charts ]{{\includegraphics[width=0.25\columnwidth]{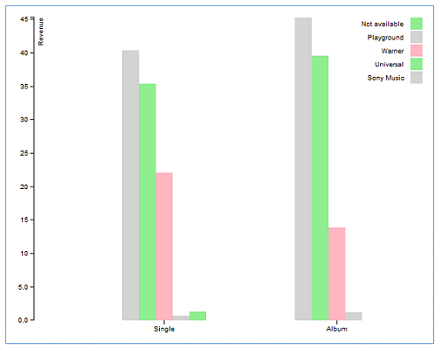}}}%
  \caption{Sample charts generated based on output of the chart recommender}%
  \label{fig:sampleCharts}%
\end{figure*}

%% file: survey2.tex
\subsection{User survey on caption-chart mappings}
\label{survey2}

One of the main assumptions in our paper is the implicit relationship between the type of items queried and the analysis task that may be performed on these data items, and hence the most relevant visual representation. We conducted a user survey to test out this hypothesis and we describe briefly the survey and the main observations. The details of methodology, and detailed findings are currently under submission to another conference. In brief, this was a crowd-sourced survey, and no assumption was made on the users' expertise with visualization, except that they all have some exposure seeing these charts in presentations at least occasionally.

We defined a questionnaire of $24$ questions, where each question contained a caption that defined the data being presented, and a suggested visualization of the data. Users were asked to rank the visualization in terms of how well it represented the caption, on a scale of $1$ to $5$, where $5$ indicated that the respondent agreed that it was a very good match and $1$ indicated a very poor match. The visualizations included both positive and negative examples in terms of best practices. In a few cases, the caption also contained keywords indicating the kind of analytical task being represented, such as  ``compare" or ``relate". In other cases these keywords were not used. $114$ participants took the survey.

The main observations are listed below.
\begin{asparaenum}
\item Overall there was a greater acceptance of the bar chart representation - where bar charts were used to visualize the data, the users broadly seemed to agree that it was acceptable.
\item In the case of positive examples, there was no perceptible difference in the respondents`` ranking of the charts, independent of whether a task-related keyword was used or not. 
\item In the case of negative examples also, there was a correlation between our ranking and the respondents` ranking with and without task-related keywords. 
\end{asparaenum}

The above seems to indicate that user expectations and acceptance of the visualization is dependent implicitly on the types of data being visualized, rather than on just the explicit task-related keywords.

%% file: futureWork.tex
\section{Conclusions and future work}
\label{futurework}

In this paper we presented the DataVizard system for generating automatic chart recommendations, based on the analysis of data and the associated metadata such as queries, schema information, table captions. We have discussed the heuristics-cased approach, and the subsequent automated approach based on feature identification and extraction, for recommending the most appropriate visualization. To the best of our knowledge, DataVizard is the first system that goes beyond recommending relatively simple charts in 2-variables. Our system can handle both SQL and non-SQL datasets, and with little programming effort can easily be embedded into any analysis workflow. Through a user-study we have shown that our approach provides significantly high-quality chart recommendation, and forms an important step towards completely automating the visual presentation of data. 

We believe that the work of recommending visual presentation is just at a very nascent stage, and there are a number of directions for future work including: expand the repertoire of visual presentations to include sophisticated visualizations; more advanced usage of data values to discover automatic aggregations that make visual presentations semantically meaningful, and more.